\pdfoutput=1

\documentclass[11pt]{article}

\usepackage[]{acl}

\usepackage{times}
\usepackage{latexsym}

\usepackage[T1]{fontenc}

\usepackage[utf8]{inputenc}

\usepackage{microtype}

\usepackage{inconsolata}

\usepackage{graphicx}
\usepackage{multirow}
\usepackage{booktabs}
\usepackage{arydshln}

\usepackage{subcaption}
\usepackage{amsmath} 

\usepackage{comment}
\usepackage{threeparttable} 

\title{Synergetic Event Understanding: A Collaborative Approach to Cross-Document Event Coreference Resolution with Large Language Models}

\author{
    Qingkai Min\textsuperscript{\rm 1,2},  
    Qipeng Guo\textsuperscript{\rm 3}, Xiangkun Hu\textsuperscript{\rm 4}, Songfang Huang\textsuperscript{\rm 5}, Zheng Zhang\textsuperscript{\rm 6}, and Yue Zhang\textsuperscript{\rm 2,7,\footnotemark[1]}
    \\
    \textsuperscript{1} Zhejiang University \ \ \ \ 
    \textsuperscript{2} School of Engineering, Westlake University \\
    \textsuperscript{3} Shanghai AI Laboratory \ \ \ \ 
    \textsuperscript{4} Fudan University \\
    \textsuperscript{5} Alibaba DAMO Academy \ \ \ \
    \textsuperscript{6} New York University Shanghai\\
    \textsuperscript{7} Institute of Advanced Technology, Westlake Institute for Advanced Study \\
    $^{2}$\texttt{\{minqingkai, zhangyue\}@westlake.edu.cn} \ \ \    
    $^{3}$\texttt{guoqipeng@pjlab.org.cn} \\
    $^{4}$\texttt{xkhu17@fudan.edu.cn} \ \ \    
    $^{5}$\texttt{songfang.hsf@alibaba-inc.com} \ \ \ \
    $^{6}$\texttt{zz@nyu.edu}
}

\begin{document}
\maketitle

\renewcommand{\thefootnote}{\fnsymbol{footnote}}
\footnotetext[1]{Corresponding author}
\renewcommand{\thefootnote}{\arabic{footnote}}

\begin{abstract}
Cross-document event coreference resolution (CDECR) involves clustering event mentions across multiple documents that refer to the same real-world events. Existing approaches utilize fine-tuning of small language models (SLMs) like BERT to address the compatibility among the contexts of event mentions. However, due to the complexity and diversity of contexts, these models are prone to learning simple co-occurrences. Recently, large language models (LLMs) like ChatGPT have demonstrated impressive contextual understanding, yet they encounter challenges in adapting to specific information extraction (IE) tasks. In this paper, we propose a collaborative approach for CDECR, leveraging the capabilities of both a universally capable LLM and a task-specific SLM. 
The collaborative strategy begins with the LLM accurately and comprehensively summarizing events through prompting. Then, the SLM refines its learning of event representations based on these insights during fine-tuning. Experimental results demonstrate that our approach surpasses the performance of both the large and small language models individually, forming a complementary advantage. Across various datasets, our approach achieves state-of-the-art performance, underscoring its effectiveness in diverse scenarios.
\end{abstract}

\section{Introduction}
Event coreference resolution is a useful task in information extraction~\citep{lu2018event}.
This is crucial for achieving a more comprehensive understanding of intricate narratives and facilitating knowledge extraction from diverse textual sources.
The coreference of events typically relies on a thorough understanding of document-level context~\citep{minh-tran-etal-2021-exploiting,kriman-ji-2021-joint,xu-etal-2022-improving}. Cross-document event coreference~\citep{lee-etal-2012-joint}, involving the comparison of event mentions from different documents, presents additional challenges. On one hand, distinct events in different documents may be portrayed in a very similar manner, especially for events of the same type (challenge 1). On the other hand, the portrayal of the identical event may vary significantly across different documents (challenge 2). The model is required to grasp comparable coreference evidence from varied contexts and make judgments based on it (refer to the examples in Table~\ref{tab:non_coref_examples} and~\ref{tab:coref_examples} for better illustration).

\begin{figure}[t]
  \centering
  \begin{subfigure}[b]{0.44\textwidth}
    \includegraphics[width=\textwidth]{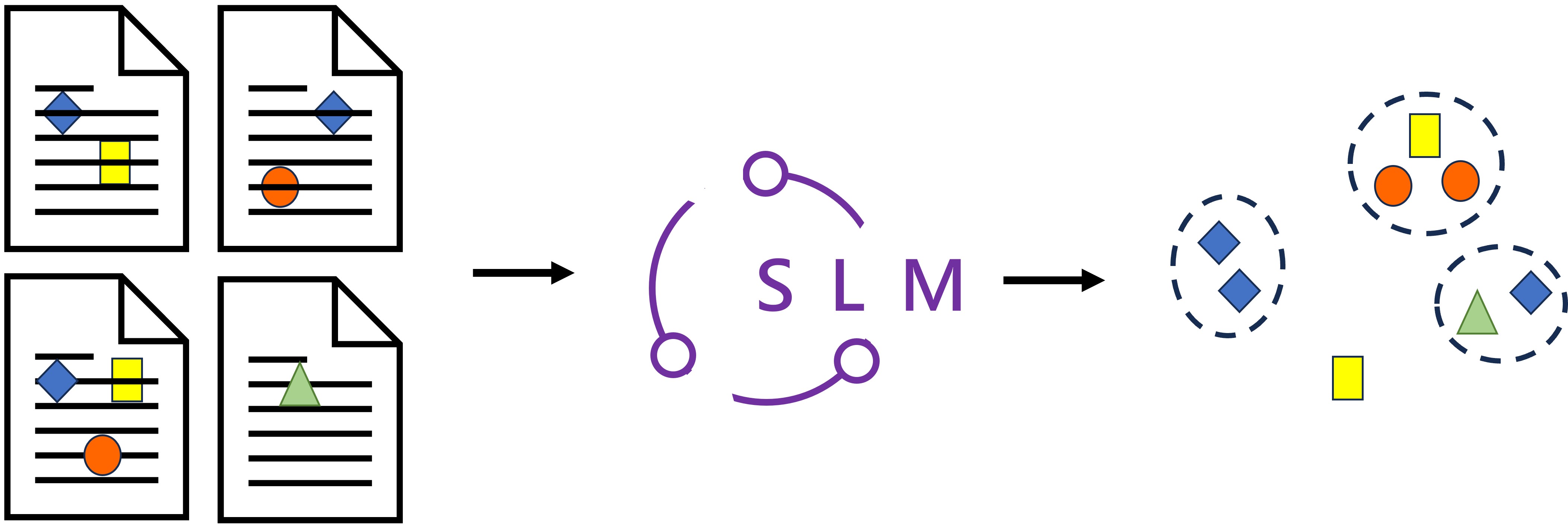}
    \caption{Existing approach}
    \label{fig:intro_slm}
  \end{subfigure}
  \hfill
  \begin{subfigure}[b]{0.44\textwidth}
    \includegraphics[width=\textwidth]{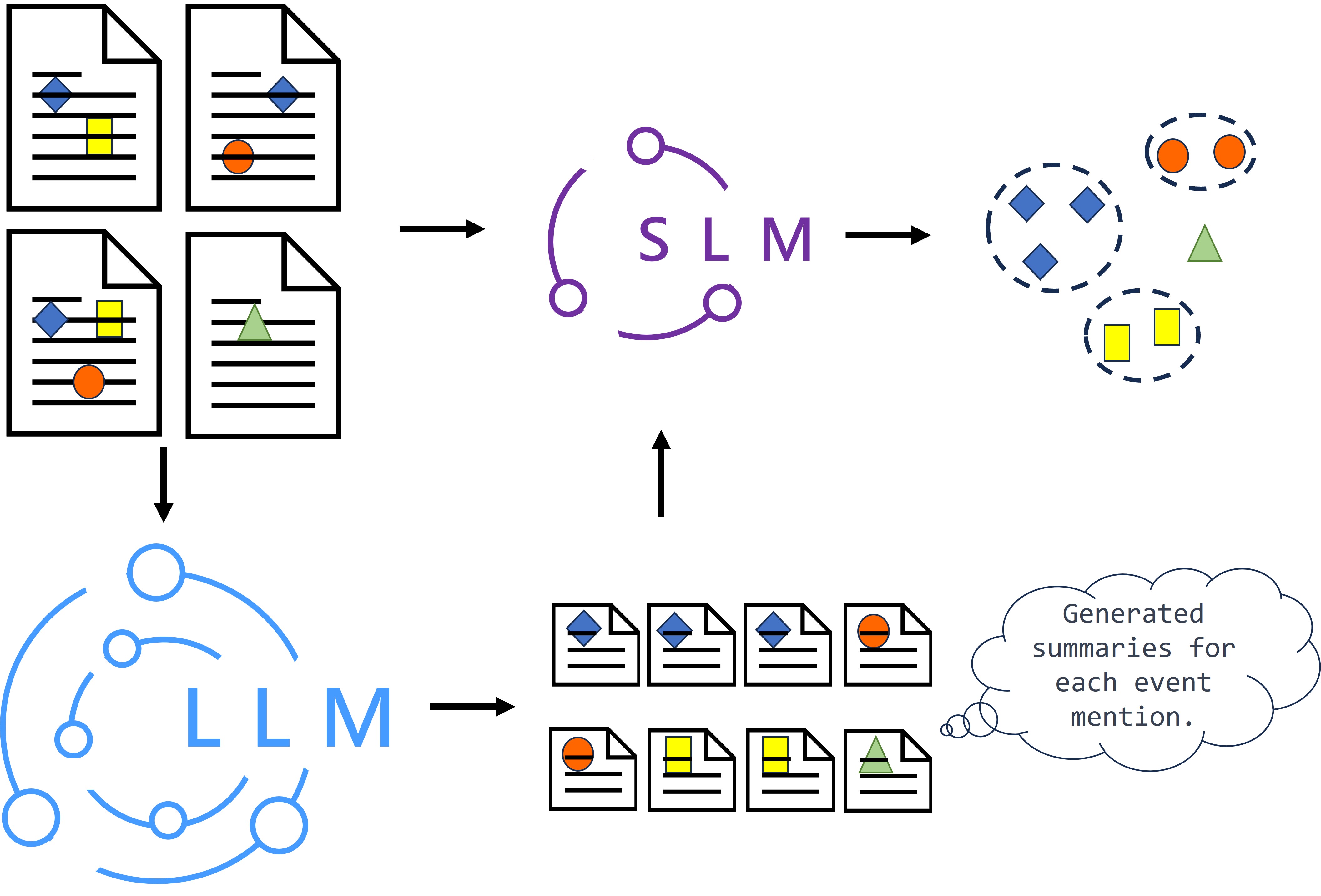}
    \caption{Our approach}
    \label{fig:intro_llm}
  \end{subfigure}
  \caption{Models for cross-document event coreference resolution, where the input comprises event mentions from different documents, and the output consists of event clusters formed by coreferential mentions, which are visually represented by icons sharing the same color and shape.}
\end{figure}

Existing work~\citep{held-etal-2021-focus,yu-etal-2022-pairwise} attempts to address CDECR based on fine-tuning small language models (SLMs)\footnote{In this work, SLM refers to pre-trained language models with relatively fewer parameters, which are more cost-effective for fine-tuning on specific tasks, such as BERT and RoBERTa.}, as shown in Figure~\ref{fig:intro_slm}. However, the complexity and diversity of the context make it prone to learning pseudo-features by capturing simple co-occurrences rather than genuinely coreference-related terms, including contextual words, entity mentions and other event mentions associated with the given event mention. 
Supporting this observation, CDECR remains a significant challenge for SLMs, as evidenced by achieving only around 70\% CoNLL F1 score on the FCC dataset~\citep{bugert-etal-2021-generalizing}.

Recent advancements in LLMs have significantly advanced the field of NLP, enabling the effective resolution for tasks like machine translation~\citep{jiao2023chatgpt} and text summarization~\citep{bang2023multitask}, with just a few demonstrations. However, when it comes to information extraction (IE) tasks, LLMs encounter challenges in task-specific adaptation. Specifically, LLMs struggle to achieve the same level of accuracy as supervised SLMs because a small number of demonstrations cannot comprehensively cover the complex annotation guidelines of these tasks~\citep{han2023information,li2023eval}. Moreover, the inherent nature of the CDECR task, which involves processing multiple documents, imposes enhanced demands on understanding the lengthy context in the demonstrations.\footnote{On average, each instance of demonstration in the ECB+ dataset contains close to 15k tokens.} 
Instead of directly predicting CDECR structures, the relative strength of LLMs can enhance the generic understanding of individual documents, particularly the inherent meaning of diverse event mentions, which is complementary to the advantage of SLMs in understanding structures with thorough fine-tuning.

To leverage the relative strengths of LLMs and SLMs, we propose a collaborative approach, as shown in Figure~\ref{fig:intro_llm}. First, we use the LLM to summarize event mentions from different documents. Then we feed these insights to the SLM to enhance its understanding of event mentions, enabling it to make coreference judgments based on more focused contexts.
For the LLM summarization, we design a two-step workflow with separate generic prompts to guide its comprehension of the context of each mention, instead of 
task-specific in-context learning or fine-tuning.
For the SLM, we employ joint representation learning to integrate the original document and the generated summary.

We conduct experiments on three datasets of CDECR, and the results demonstrate that our collaborative approach, as compared to methods solely relying on the LLM or SLM, exhibits significant improvements. Across all three datasets (ECB+, GVC, and FCC), our approach achieves state-of-the-art results, with increases of 1\%, 2.7\%, and 7\% in CoNLL F1, respectively (averaged over three independent experiments for each dataset). Through analysis, it is demonstrated that our approach more thoroughly addresses the aforementioned challenge 1 of similarly portrayed contexts, making a substantial contribution to performance improvement.

To the best of our knowledge, we are the first to propose a collaborative approach that leverages the universal capabilities of LLMs to address CDECR, achieving superior performance compared to the state-of-the-art baseline.\footnote{The code and data are publicly available at \href{https://github.com/taolusi/SECURE}{https://github.com/taolusi/SECURE}.}

\section{Related Work}

\textbf{CDECR} Early work addresses CDECR by employing machine learning methods with manually designed features~\citep{bejan-harabagiu-2010-unsupervised,yang-etal-2015-hierarchical,vossen2018identity,bugert-etal-2021-generalizing}. Recent neural approaches have utilized SLMs to encode event mentions, obtaining their embeddings for supervised coreference resolution. 
Initial efforts involve encoding at sentence level and fusing the embeddings of mentions and the incomplete arguments extracted by SRL as the representation of mentions~\citep{barhom-etal-2019-revisiting,zeng-etal-2020-event,allaway-etal-2021-sequential,yu-etal-2022-pairwise}. Subsequent work incorporates extensive context directly into encoding, leading to noticeable improvements~\citep{caciularu-etal-2021-cdlm-cross,cattan-etal-2021-cross-document,held-etal-2021-focus,hsu-horwood-2022-contrastive,ahmed-etal-2023-2}. 
More recently, \citet{chen-etal-2023-cross} and \citet{ravi-etal-2023-happens} establish connections between event mentions using a discourse rhetorical structure constructor and a GPT-3 model fine-tuned with additional data for temporal reasoning, respectively. In comparison to existing work, we are the first to establish comprehensive connections between event mentions and their corresponding contextual elements, including contextual words, entity mentions, and other event mentions, by leveraging the intrinsic knowledge and out-of-the-box context comprehension ability of LLMs.

\textbf{LLM for IE} Several recent studies~\citep{ma-etal-2023-large,li2023eval,han2023information,yuan-etal-2023-zero,gao2023exploring,wei2023zero,xie-etal-2023-empirical,li2023far,xu-etal-2023-unleash,wadhwa-etal-2023-revisiting,qi-etal-2023-preserving,ling2023improving} have evaluated the performance of LLMs, predominantly ChatGPT, using in-context learning methods on various IE tasks.
These investigations universally demonstrate that LLMs exhibit commendable performance in zero-shot and few-shot settings, yet there remains a substantial gap when compared to state-of-the-art supervised SLMs, with the performance gap widening for more complex tasks.

In addition, there are also methods directly using labeled data from IE tasks to fine-tune LLMs\citep{lu-etal-2022-unified,zhou2023universalner,wang2023instructuie,sainz2023gollie}. In general, training on these larger-scale models, such as Code-LLaMA and Flan-T5, has yielded results comparable to supervised baselines and demonstrated improvements in zero-shot settings. However, when the training of LLMs does not result in significant performance gains, the training cost, compared to training SLMs, becomes less cost-effective.

\textbf{Integration of LLM and SLM}
The integration of LLM and SLM is an emerging approach, with only a few explorations in complex IE tasks. \citet{ma-etal-2023-large} prompts the LLM to rerank a few difficult samples filtered by the supervised SLM and achieves improvements on various few-shot IE tasks. Their method is based on the observation that LLMs excel only at a small number of hard samples.
\citet{wan-etal-2023-gpt} first utilizes the LLM to generate reasoning logic for demonstrations retrieved by a fine-tuned SLM, then feeds this combined input back to the LLM for relation extraction, surpassing supervised baselines on some datasets.
An inherent challenge lies in finding reasonable demonstrations for NULL-type triples, leading to poor performance on complex tasks such as ACE05. Additionally, inducing complex reasoning logic for each of the k-demonstrations is costly, leading them to sample only a subset of ACE05 and TACRED test sets.
\citet{xu-etal-2023-unleash} and \citet{li-etal-2023-semi} leverage LLMs for data enhancement in sentence and document-level relation extraction tasks, respectively.
The gap between triples recognized by LLMs and those annotated under manually crafted rules introduces potential shifts in data distribution, making the effectiveness in applications unclear.

Overall, the aforementioned integration methods have exhaustively attempted to adapt LLMs to specific tasks by prompting them to establish accurate connections with artificially defined labels.
In contrast, our approach only requires LLMs to perform generic tasks, leveraging their inherent capabilities to assist specific tasks.

Concurrently, akin to our approach, \citet{ding2024rationale} and \citet{nath2024okay} also leverage LLM generation to assist SLM on CDECR.
While \citet{ding2024rationale} prompts LLM with task instructions to generate multiple counterfactual instances for original mention pairs, \citet{nath2024okay} employs similar task prompts to guide LLM in generating coreference reasoning processes for mention pairs. Unlike our approach, which involves a general task of having LLM process each mention individually, their methods require LLM to directly handle the relationships between mentions given coreference labels.
In terms of efficiency, their methods are less effective than ours as they need to handle combinations of mention pairs, resulting in a quadratic increase in the number of processing entries.

\begin{table*}[!ht]
\centering
\resizebox{\textwidth}{!}{%
\fontsize{6}{7}\selectfont
\begin{tabular}{ll}
\hline
Step 1 &
  \begin{tabular}[c]{@{}l@{}}\textbf{News}: \texttt{[input document]}\\ \textbf{Question 1:} In this news, given ``\texttt{[mention 1]}'' mentioned in the sentence ``\texttt{[the sentence]}''.\\ Please elaborate \texttt{[mention 1]} in the context of the news article. \\ Present the information in the following format: `Elaboration: \texttt{[mention 1]} refers to <placeholder>'. \\ \textbf{Question 2:} ...
  \end{tabular}\\ \hline
Step 2 &
  \begin{tabular}[c]{@{}l@{}}\textbf{News:} \texttt{[input document]}\\ \textbf{Question 1:} In this news, given ``\texttt{[mention 1]}'' mentioned in the sentence ``\texttt{[the sentence]}''.\\ Elaboration: \texttt{[output from step 1]}. \\Please further elaborate ``\texttt{[mention 1]}'' by providing details for entities in the elaboration utilizing\\ coreference resolution. Provide any available or approximate dates in the news for reference, which\\ can be inferred from the publication date of the news if available. \\ Present the information in the following format: `Elaboration: \texttt{[mention 1]} refers to <placeholder>'.  \\ \textbf{Question 2:} ...
  \end{tabular} \\ \hline
\end{tabular}%
}
\caption{The two-step workflow for LLM summarization. Each prompt includes a document along with multiple event mentions. 
Step 2 takes the output from Step 1 as its input.
The content to be filled is represented as [\texttt{content}].} 
\label{tab:workflow}
\end{table*}

\section{Method}
We adopt the state-of-the-art method proposed by~\citet{held-etal-2021-focus} as our baseline (Section~\ref{sec:baseline}), then summarize events using generic prompts for LLM (Section~\ref{sec:llm_sum}), and finally integrate the representations of events from both the summary and the original context into baseline system (Section~\ref{sec:slm_inte}).

\subsection{Task and Baseline}
\label{sec:baseline}
The goal of the CDECR task is to group coreferential event mentions across multiple documents into clusters. We formalize the task as follows:

\textbf{Input:} A corpus comprising multiple documents denoted by \(D\), where \(D = \{D_1, D_2, ..., D_{|D|}\}\), with \(|D|\) representing the number of documents in the dataset. Let \(M\) represent all event mentions in the corpus, such that \(M = \{m_{11}, m_{12}, \ldots, m_{ij}, \ldots, m_{|D|, k}\}\), where \(k\) denotes the number of event mentions in each document, and \(m_{ij}\) signifies the \(j\)-th event mention in document \(D_i\).

\textbf{Output:} A set of clusters, denoted as \(C\), where \(C = \{C_1, C_2, ..., C_n\}\).
For each cluster \(C_k\), \(E_k\) represents all the event mentions contained in the cluster \(C_k\), such that \(E_k = \{e_{k1}, e_{k2}, \ldots, e_{kj}, \ldots, e_{kM}\}\), where \(M\) is the total number of event mentions in cluster \(C_k\), and \(e_{kj}\) is the \(j\)th event mention in the cluster \(C_k\).

Our baseline consists of two key modules for coreference clustering: candidate retrieval and pairwise classification. Both of these modules primarily involve using a RoBERTa~\citep{liu2019roberta} encoder to encode the context and obtain vector representations of event mentions, which can be seamlessly replaced by our collaborative approach. We formalize the encoding process as follows:

For each event mention \(m_{ij}\), its vector representation can be obtained as:
\begin{equation}
h_{ij} = f_{\text{enc}}(m_{ij}, D_i)
\end{equation}
Here \(f_{\text{enc}}\) is an encoder network used for encoding \(D_i\) and concatenating the representations of the boundary tokens of \(m_{ij}\). The resulting representation \(h_{ij}\) is fed into the subsequent neural network.

\subsection{LLM Summarization}
\label{sec:llm_sum}

Summarizing events for CDECR poses a non-trivial challenge. Existing summarization methods are typically designed to provide a general overview of documents, making it difficult to extract information specific to certain types of events. This not only provides limited assistance for coreference but may also lead to the omission of crucial details. 
Furthermore, designing a summary template for each type of event is not only impractical in real-world applications\footnote{Based on the rough statistics in our experiment, the ECB+ dataset contains over 400 event types.} but also introduces bias, potentially causing LLMs to misinterpret or hallucinate information due to the inherent incompleteness of event information in documents.

To address various types of events and gather specific details from complex contexts, we design a two-step workflow to prompt the LLM, as shown in Table~\ref{tab:workflow}. The first step is responsible for extracting tailored information for different types of events in the context of the document.
The second step aims to expand the details of the entities mentioned in the output of the first step, as entity details are often scattered throughout the document.
In each step, we employ a straightforward prompt to accomplish a primary task. Our prompts adhere to the basic principle of faithfulness, avoiding additional interpretations to prevent semantic shifts. Compared to a synthesized single-step workflow, our two-step workflow guarantees that each step remains focused on its main objective, thereby preventing interference between the two steps, as illustrated by the analysis in Section~\ref{sec:anal_workflow}.

In the first step, we instruct the LLM agent to ``elaborate'' an event mention, rather than the conventional instruction of ``summarize''.
The term ``elaborate'' implies an explanatory behavior based on the concept of the mention words themselves, emphasizing the support of details from the document context. This suggests that LLMs can automatically select any relevant details from the context to support this explanation, including contextual words, entity mentions, and event mentions.
This provides a standardized and feasible way to understand events, leveraging the LLM's intrinsic knowledge and contextual understanding capabilities without imposing complex rules for the LLM to adhere to. 

In the second step, we prompt the LLM agent to use coreference resolution to aggregate detailed information about entities, as entity coreference is a more standardized task compared with event and performing it within a document reduces complexity. Additionally, we require the LLM to perform temporal reasoning based on the publication date of the document, further reducing ambiguity in coreference evidence comparison.

In both steps, we specify the generation format to ensure the consistency between the mention spans in summary and original document.  
This not only reduces the generation difficulty of LLM but also facilitates SLM in establishing the connection between the two during joint representation learning.

\subsection{Integration into Final SLM}
\label{sec:slm_inte}
The SLM takes the original document and the generated summary as inputs. Through a direct joint representation learning technique, the new mention vector representation can be seamlessly integrated into the baseline.

Specifically, for the mention \(m_{ij}\), let \(S_{ij}\) represents the generated summary, and \(m_{ij}^{(s)}\) signifies the mention within it. By concatenating the original document \(D_i\) and the summary \(S_{ij}\), a new document \(D_i'\) is formed. Let \(f_{\text{enc}}'\) denotes the new encoder network. It first encodes the new document \(D_i'\), obtaining vector representations \(h_{ij}\) and \(h_{ij}^{(s)}\) for \(m_{ij}\) and \(m_{ij}^{(s)}\) respectively. These vectors are then concatenated to form the fused mention vector representation \(h_{ij}'\), which can seamlessly substitute \(h_{ij}\) in the baseline for subsequent operations. The joint representation learning process can be represented as:
\begin{align*}
h_{ij}' & = f_{\text{enc}}'(\{e_{ij}, e_{ij}^{(s)}\}, D_i') \\
& = \text{concat}\left(f_{\text{enc}}(\{e_{ij}, e_{ij}^{(s)}\}, D_i')\right) \\
& = \text{concat}(h_{ij}, h_{ij}^{(s)})
\end{align*}
Here \(\{e_{ij}, e_{ij}^{(s)}\}\) denotes a set containing two elements, implying that vector representations for both \(e_{ij}\) and \(e_{ij}^{(s)}\) can be derived using the same process as for a single element.

This integration method, which involves concatenating the original context and generated summary for joint representation learning, enables mutual learning of each other's context in the same attention space, thereby enhancing the understanding of genuinely coreference-related terms.

\begin{table*}[!ht]
\centering
\resizebox{\textwidth}{!}{%
\begin{tabular}{lccccccccccccc}
\hline
\multirow{2}{*}{Methods} & \multicolumn{3}{c}{MUC} & \multicolumn{3}{c}{$B^3$} & \multicolumn{3}{c}{CEAF$_e$} & CoNLL & \multicolumn{3}{c}{LEA} \\ \cline{2-14} 
 & R & P & F1 & R & P & F1 & R & P & F1 & F1 & R & P & F1 \\ \hline
\textbf{ECB+} &  &  &  &  &  &  &  &  &  &  &  &  &  \\
\citet{barhom-etal-2019-revisiting} & 77.6 & 84.5 & 80.9 & 76.1 & 85.1 & 80.3 & 81.0 & 73.8 & 77.3 & 79.5 & - & - & - \\
\citet{cattan2020streamlining} & 85.1 & 81.9 & 83.5 & 82.1 & 82.7 & 82.4 & 75.2 & 78.9 & 77.0 & 81.0 & - & - & - \\
\citet{bugert-etal-2021-generalizing} & 76.0 & 76.1 & 76.1 & 71.8 & 81.2 & 76.2 & 72.2 & 72.1 & 72.2 & 74.8 & 55.1 & 67.9 & 60.8 \\
\citet{caciularu-etal-2021-cdlm-cross} & 87.1 & \textbf{89.2} & 88.1 & 84.9 & \textbf{87.9} & 86.4 & \textbf{83.3} & 81.2 & 82.2 & 85.6 & 76.7 & 77.2 & 76.9 \\
\citet{held-etal-2021-focus} & 87.0 & 88.1 & 87.5 & 85.6 & 87.7 & 86.6 & 80.3 & \textbf{85.8} & 82.9 & 85.7 & 74.9 & 73.2 & 74.0 \\
\citet{hsu-horwood-2022-contrastive} & 87.8 & 82.9 & 85.3 & 86.5 & 83.1 & 84.8 & 76.9 & 82.8 & 79.7 & 83.3 & 74.4 & 74.0 & 74.2 \\
\citet{yu-etal-2022-pairwise} & 88.1 & 85.1 & 86.6 & 86.1 & 84.7 & 85.4 & 79.6 & 83.1 & 81.3 & 84.4 & - & - & - \\
\citet{ahmed-etal-2023-2}\footnotemark[6] & 80.0 & 87.3 & 83.5 & 79.6 & 85.4 & 82.4 & 83.1 & 75.5 & 79.1 & 81.7 & 70.5 & 73.3 & 71.9 \\
\citet{chen-etal-2023-cross} & 88.6 & 85.9 & 87.2 & 87.8 & 85.4 & 86.6 & 82.8 & 83.7 & 83.2 & 85.7 & - & - & - \\
GPT-4 & 79.8 & 78.0 & 78.9 & 76.3 & 78.1 & 77.2 & 73.3 & 75.6 & 74.4 & 76.8 & 65.0 & 70.0 & 67.4 \\
Our baseline & 86.6 & 86.8 & 86.7 & 87.1 & 86.0 & 86.5 & 82.6 & 82.5 & 82.5 & 85.2 & 77.8 & 76.6 & 77.2 \\
Our method & \textbf{89.4} & 87.1 & \textbf{88.2} & \textbf{89.1} & 86.5 & \textbf{87.8} & 82.7 & 85.5 & \textbf{84.1} & \textbf{86.7} & \textbf{79.7} & \textbf{78.5} & \textbf{79.3} \\ \hline
\textbf{GVC} &  &  &  &  &  &  &  &  &  &  &  &  &  \\
\citet{barhom-etal-2019-revisiting} & - & - & - & 81.0 & 66.0 & 72.7 & - & - & - & - & - & - & - \\
\citet{bugert-etal-2021-generalizing} & 66.3 & 78.1 & 71.7 & 49.9 & 73.6 & 59.5 & 60.9 & 38.2 & 47.0 & 59.4 & 38.2 & 56.5 & 45.6 \\
\citet{held-etal-2021-focus} & 91.8 & 91.2 & 91.5 & 82.2 & 83.8 & 83.0 & 75.5 & 77.9 & 76.7 & 83.7 & 79.0 & 82.3 & 80.6 \\
\citet{ahmed-etal-2023-2} & 84.0 & 91.1 & 87.4 & 79.0 & 76.4 & 77.7 & 69.6 & 52.5 & 59.9 & 75.0 & 74.1 & 63.9 & 68.6 \\
GPT-4 & 7.6 & 54.9 & 13.4 & 5.5 & 34.6 & 9.6 & 4.2 & 42.8 & 7.6 & 10.2 & 4.4 & 28.0 & 7.6 \\
Our baseline & 91.3 & 92.0 & 91.7 & 86.2 & 83.8 & 84.9 & 78.7 & 76.5 & 77.6 & 84.7 & 82.0 & 78.4 & 80.2 \\
Our method & \textbf{92.4} & \textbf{93.2} & \textbf{92.8} & \textbf{87.0} & \textbf{87.4} & \textbf{87.2} & \textbf{83.6} & \textbf{80.7} & \textbf{82.1} & \textbf{87.4} & \textbf{83.4} & \textbf{83.0} & \textbf{83.2} \\ \hline
\textbf{FCC} &  &  &  &  &  &  &  &  &  &  &  &  &  \\
\citet{barhom-etal-2019-revisiting} & - & - & - & 36.0 & \textbf{83.0} & 50.2 & - & - & - & - & - & - & - \\
\citet{bugert-etal-2021-generalizing} & 82.7 & 78.3 & 80.4 & 70.8 & 38.3 & 49.2 & 28.2 & 40.4 & 33.2 & 54.3 & 60.4 & 30.4 & 39.8 \\
\citet{held-etal-2021-focus} & \textbf{86.4} & 75.7 & 80.7 & 61.6 & 65.4 & 63.5 & 39.1 & \textbf{65.3} & 48.9 & 64.4 & 47.2 & 57.0 & 51.6 \\
GPT-4 & 0.1 & 1.0 & 0.2 & 2.3 & 99.4 & 4.5 & 14.1 & 13.1 & 13.6 & 6.1 & 0.0 & 1.1 & 0.0 \\
Our baseline & 81.4 & 89.0 & 85.1 & 69.4 & 66.6 & 68.0 & 76.4 & 52.2 & 62.0 & 71.7 & 63.5 & 54.6 & 58.7 \\
Our method & 85.3 & \textbf{90.6} & \textbf{87.8} & \textbf{74.5} & 82.5 & \textbf{78.3} & \textbf{80.9} & 61.5 & \textbf{69.8} & \textbf{78.7} & \textbf{69.7} & \textbf{73.5} & \textbf{71.5} \\ \hline
\end{tabular}%
}
\caption{Performance comparison on the ECB+, GVC, and FCC datasets. Our baseline results are obtained by replicating the state-of-the-art method proposed by Held et al. (2021), with the adoption of more advanced hyper-parameters. 
Our method shows a statistically significant improvement compared to our baseline, with a significance level of \(p < 0.01\). The results of GPT-4 are based on the best-performing method, specifically through few-shot learning with limited context. The best results are highlighted in bold.}
\label{tab:main_result}
\end{table*}

\section{Experiments}
\subsection{Experimental Settings}
\label{sec:setting}
\paragraph{Dataset}
We conduct experiments on three CDECR datasets: Event Coreference Bank Plus (ECB+)~\citep{cybulska-vossen-2014-using}, Gun Violence Corpus (GVC)~\citep{vossen-etal-2018-dont}, and Football Coreference Corpus (FCC)~\citep{bugert-etal-2021-generalizing}.
The widely-used ECB+ dataset consists of news articles from various topics, including earthquakes, murders, acquisitions, etc. Each topic includes two similar subtopics, such as ``6.1 earthquake Indonesia 2009'' and ``6.1 earthquake Indonesia 2013''.
This setup aligns with the challenge 1 mentioned in introduction, asking the model to distinguish similar events. Similarly, GVC and FCC, focusing on news incidents of gun violence and football tournaments, respectively, also have multiple subtopics under one overarching topic. More details can be found in Table~\ref{tab:dataset_stat} (Appendix~\ref{sec:dataset}).

\paragraph{Evaluation Metrics} Following previous work~\citep{barhom-etal-2019-revisiting,held-etal-2021-focus}, we conduct a comprehensive comparison using metrics including MUC, $B^3$, CEAF$_e$, CoNLL, and LEA. The CoNLL F1 is a composite metric representing the average of the first three. $B^3$ is chosen for analysis, following~\citet{held-etal-2021-focus}.
\paragraph{Hyper Parameters} 
For LLM summarization, we use the ``\texttt{GPT-4-0613}'' model via OpenAI API, setting the sampling temperature \(t=0\) to reduce the impact of randomness. In the first step of the generation workflow, we introduce a pre-step of instructing the LLM to perform dependency parsing on the sentence containing the event mention. Based on the parsing results, the LLM then elaborates on the mention.
For SLM integration, we employ the pre-trained RoBERTa$_{\text{\scriptsize LARGE}}$ model~\citep{liu2019roberta} to embed event mentions, following our baseline~\citep{held-etal-2021-focus}. For all three datasets, we apply a consistent set of hyper-parameters for fine-tuning, as detailed in Table \ref{tab:fine_tune_param} (Appendix~\ref{sec:hyper_param}). In all experiments, be it primary results or analyses, we ensure reliability by conducting three independent experiments and averaging the outcomes.

\paragraph{Directly Using LLM to Predict the Structure of CDECR} 
We test the performance of GPT-4 using different in-context learning methods, including few-shot and zero-shot learning, with varied contexts, such as full context and mention-inclusive sentences\footnote{``Mention-inclusive sentences'' indicates that we retain only those sentences containing mentions, reducing the complexity of contextual understanding.}. 
The inherent nature of the CDECR task poses a challenge for LLMs in dealing with inputs (comprising hundreds of documents as context) and outputs (consisting of coreference structures formed by thousands of event mentions) that exceed manageable lengths.
To tackle this, we first opt for the ``\texttt{GPT-4-Turbo-Preview}'' model from OpenAI, which supports input up to 120k tokens and output  to 4096 tokens.
Second, we partition the data by topic and process it sequentially. 
Each time, all documents within a single topic are used as input, and the outputs from all topics are simply merged for testing. 
Note that this procedure only applies to the multi-topic ECB+ dataset, as there are no cross-topic links. For the single-topic GVC and FCC datasets, inputs and outputs exceeding the length limit are directly truncated.
More implementation details, including prompt design as well as the selection and parameter settings of the GPT-4 models, can be found in Appendix~\ref{sec:gpt_prompt}.

\footnotetext[6]{For fairness, results obtained under their custom oracle setting, which utilizes gold coreference information from the dev and test sets, are not included in the comparison.}
\addtocounter{footnote}{1}

\subsection{Results}
The main results are presented in Table~\ref{tab:main_result}.
Our method achieves new state-of-the-art results on all three datasets, outperforming both the previously reported best results and the improved results obtained by our reproduced baseline.

\textbf{ECB+} On this widely studied dataset, our method demonstrates improvements of 1.5\% in CoNLL F1, compared to our baseline. In comparison to~\citet{held-etal-2021-focus}, upon which our baseline is built, our method also exhibits a 1\% increase in CoNLL F1.\footnote{Our baseline is slight lower in CoNLL F1 than reported by \citet{held-etal-2021-focus}, potentially be attributed to randomness.} This improvement stands out notably in recent research,
accompanied by significance testing to demonstrate its robustness.
Compared to \citet{chen-etal-2023-cross}, who also employs RoBERTa$_{\text{\scriptsize LARGE}}$ for encoding while proposing a different method to leverage broader contexts, we also achieve a 1\% improvement, showcasing the effectiveness of our method in utilizing context.
More experiments and discussions under additional evaluation principles, including without singletons and at the topic level, can be found in the Appendix~\ref{sec:eval_singleton}.

GPT-4 utilizing few-shot learning significantly lags behind our method, with nearly a 10\% gap in CoNLL F1, indicating that GPT-4 still faces substantial adaptability challenges in directly predicting cross-document event coreference structures. This also demonstrates the effectiveness of our method in leveraging the inherent general capabilities of LLMs.
Further analysis of GPT-4's performance can be found in Section~\ref{sec:anal_gpt4}. We also compare the efficiency of LLM utilization (in terms of number of API calls and token consumption) between summarization and structure prediction, as detailed in Appendix~\ref{sec:effi_trunc}.

\textbf{GVC \& FCC}
Our method demonstrates improvements of 2.7\% and 7.0\% in CoNLL F1 on the GVC and FCC datasets, respectively, compared to our baseline. The significant improvement on the challenging FCC dataset further underscores the effectiveness of our collaborative approach in leveraging LLM. Additionally, our baseline also shows improvements of 1.0\% and 7.3\% compared to \citet{held-etal-2021-focus}, highlighting our comprehensive exploration on these two less-studied datasets.

GPT-4 exhibits abnormal performance on the GVC and FCC datasets, primarily due to truncation issues stemming from its length constraints, as mentioned in Section~\ref{sec:setting}. This is more pronounced on FCC, where longer multi-document contexts are encountered in the test set compared to GVC (4274 vs 1360 sentences).
Further elaboration on the truncation problem can be found in Appendix~\ref{sec:effi_trunc}.

\subsection{The Impact of LLM Summarization}
\paragraph{Error Analysis}
\label{sec:error_anal}
To gain a deeper understanding of the improvements achieved through LLM summarization, we perform a quantitative analysis on the false links within the clusters (see Table~\ref{tab:error_stat}). 

Similar to \citet{yu-etal-2022-pairwise}, we categorize link errors into two types: false positive (FP) and false negative (FN). FP links (incorrect links) occur when two non-coreferential mentions are clustered together, while FN links (missing links) occur when two coreferential mentions are not clustered together.
Additionally, we further categorize FP links into two sub-types based on whether two mentions share the same event type.\footnote{Details are provided in Appendix~\ref{sec:type_cate}.}
FPA (false positives caused by arguments) indicates that two mentions of the same type differ in argument information. FPT (false positives caused by types) implies that two mentions actually belong to different event types, eliminating the need to consider arguments.

\begin{table}[t]
\centering
\resizebox{0.4\textwidth}{!}{%
\begin{tabular}{llccc}
\hline
\multicolumn{1}{l}{Dataset} & \multicolumn{1}{l}{Method} & FPA   & FPT & FN \\ \hline
\multirow{2}{*}{ECB+}       & Our baseline                   & 1775  & 302 & 1262 \\ 
& Our method                  & 1227  & 152 & 1087 \\ \hline 
\multirow{2}{*}{GVC}        & Our baseline                   & 1412  & 13  & 1041 \\ 
& Our method                  & 865  & 13  & 1173 \\ \hline 
\multirow{2}{*}{FCC}        & Our baseline                   & 38522 & 0   & 8978 \\ 
& Our method                  & 4037  & 20  & 8575 \\ \hline
\end{tabular}%
}
\caption{Statistics of errors by different types.}
\label{tab:error_stat}
\end{table}

\begin{table*}[ht]
\centering
\resizebox{\textwidth}{!}{
\fontsize{10}{12}\selectfont
\begin{tabular}{p{0.63\textwidth}p{0.37\textwidth}}
\hline
\multicolumn{1}{l}{Context} &
\multicolumn{1}{l}{Summarization} \\ \hline
Dozens injured, child dead as [\textit{6.1 - magnitude earthquake}] hits Indonesia's Aceh Updated : July 02, 2013 15:50 IST A 6.1-magnitude earthquake which hit the Indonesian province of Aceh on Tuesday killed a child, injured dozens and destroyed buildings ... The quake struck inland at 0737 GMT at a depth of just 10 kilometres (6.2 miles) ... the US Geological Survey said . House collapsed ... 50 people with injuries ... 30 people seriously injured ... People panicked and rushed out of their homes ... In 2004 a massive tremor sparked a tsunami ... &
[\textit{6.1-magnitude earthquake}] refers to the seismic event that occurred in the \textbf{Bener Meriah district in the heart of Aceh}, Indonesia, on \textbf{July 2, 2013}. The earthquake struck inland at 0737 GMT at a depth of just 10 kilometres (6.2 miles) and was felt strongly for around 15 seconds, \textbf{from Bener Meriah to Banda Aceh}. \\ \hline
Indonesia's West Papua province was hit by a magnitude 6.1 [\textit{earthquake}] today, the latest powerful tremor to shake the region where five people were killed and hundreds injured at the weekend when buildings were destroyed. The quake struck off the coast at 7:48 a.m. local time, 75 kilometers (50 miles) ... the U.S. Geological Survey said ... At least five people were killed, 250 others injured and more than 800 homes destroyed ... 14,000 people fled their homes ... temblor in 2004 caused a tsunami ... &
[\textit{earthquake}] refers to the magnitude 6.1 earthquake that hit Indonesia's \textbf{West Papua} province on an \textbf{unspecified date}. The earthquake struck off the coast at 7:48 a.m. local time, 75 kilometers (50 miles) west of the region's \textbf{main city of Manokwari}, according to the U.S. Geological Survey. \\ \hline
\end{tabular}
}
\caption{Two non-coreferential mentions for the event type ``earthquake'', illustrating the remarkably similar contexts, as well as our generated more distinctive summaries. To better illustrate the similarity, we preserve the sentence containing the mention along with similar content from the context. Key information in our summarization is highlighted in bold. Mention spans are represented as [\textit{mention span}].}
\label{tab:non_coref_examples}
\end{table*}

\textbf {FPA} 
Our method demonstrates the most substantial reduction in FPA errors across all three datasets, making the greatest contribution to the overall improvement. The reduction is approximately 30\% for both ECB+ and GVC, and nearly 90\% for FCC.
The significant reduction on FCC is primary attributed to its nature, comprising multiple consecutive events from a large tournament, resulting in more pronounced contextual similarities.
This underscores the effectiveness of our method in distinguishing events with similar contextual narratives (aligning with the challenge 1 from introduction).
In Table~\ref{tab:non_coref_examples}, we present instances illustrating two highly similar earthquakes. The original context includes details about the earthquake occurrence, earthquake casualties, media coverage, and historical events. Our generated summaries primarily focus on the core details of the earthquakes, such as date and specific location, thus facilitating their differentiation.
It can be observed that our LLM summarization is capable of identifying specific information for particular events and aggregating sufficient details from the entire context.

\textbf{FPT} 
Compared to FPA, there are significantly fewer FPT errors, only appearing in the ECB+ dataset. The few occurrences on the single-topic GVC and FCC datasets can likely be disregarded, possibly due to random factors. This is because in the multi-topic ECB+ dataset, there may be topics in the test set that were not encountered in the training set, leading to unseen event types. By reducing half of the FPT errors on the ECB+ dataset, it signifies that our summarization also assists in distinguishing unseen event types.

\textbf{FN}
Our method shows less improvement in reducing FN errors compared to FP. The challenges arise from two primary factors. 
Firstly, mentions of the same event can vary greatly in expression styles.
Secondly, some event mentions naturally lack sufficient details as the authors assume that readers already possess necessary background information. 
We illustrate these issues with instances in Appendix~\ref{sec:fn_instances}.
For these cases (aligning with the challenge 2 in introduction), additional training data or external information retrieval may be necessary, as our faithful summarization based on the original context struggles to cope.

Overall, LLM summarization excels in consolidating information for specific events, facilitating the differentiation of similar yet non-coreferential events. Relatively, its effectiveness is limited for events with significant expression differences or those lacking essential details.

\paragraph{LLM Summarization VS LLM Paraphrase}

\begin{figure}[t]
    \centering
    \includegraphics[width=0.9\linewidth]{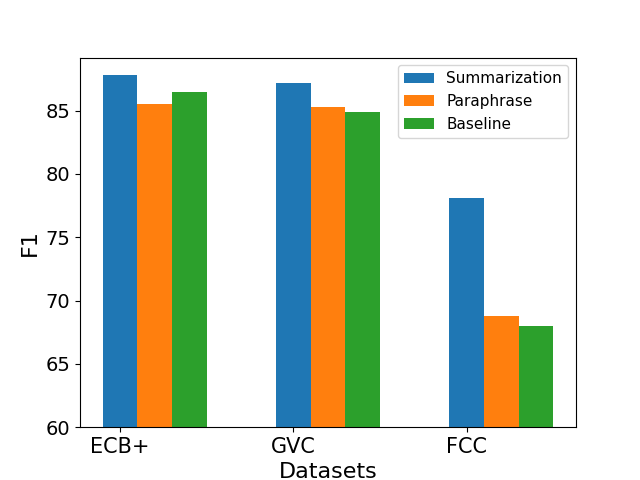}
    \caption{LLM paraphrase comparison with $B^3$ F1. The vertical axis has a baseline starting from 60.}
    \label{fig:paraphrase_f1}
\end{figure}

To validate that the performance improvement brought by our summarization is due to genuinely extracting crucial information rather than introducing diversity in context, we conduct a  comparison with paraphrases generated by the LLM. We prompt the LLM to paraphrase the context of mentions instead of the sentences they belong to, and use the same hyper-parameters for fine-tuning the SLM. As shown in Figure~\ref{fig:paraphrase_f1}, compared to our baseline, LLM paraphrase exhibits a slight improvement on GVC and FCC, with a more pronounced decline on ECB+. More importantly, it significantly lags behind our summarization on all datasets. This comparison demonstrates the capability of our summarization method in selecting and aggregating relevant information. The prompt for LLM paraphrase is provided in Table~\ref{tab:para_prompt} (Appendix~\ref{sec:para_prompt}).

\subsection{Ablation Study on the Two-step Workflow}
\label{sec:anal_workflow}

\begin{figure}[t]
    \centering
    \includegraphics[width=0.9\linewidth]{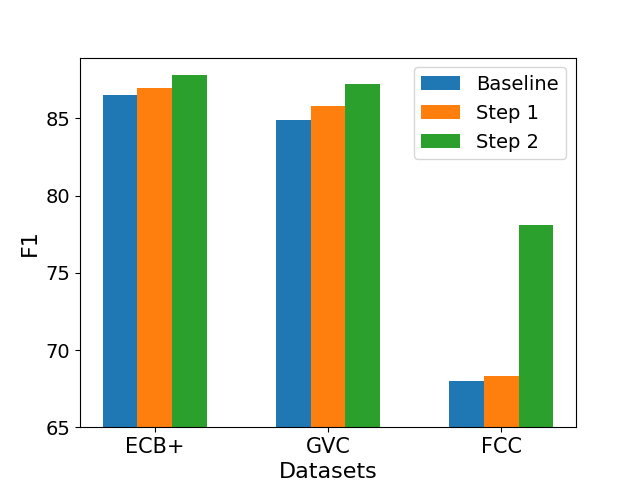}
    \caption{Comparison of different steps with $B^3$ F1. The vertical axis has a baseline starting from 65.}
    \label{fig:multi-step_f1}
\end{figure}

We conduct an ablation study to specifically illustrate the effect of Step 1 and Step 2 in LLM summarization (Table~\ref{tab:workflow}). As shown in Figure~\ref{fig:multi-step_f1}, both steps contribute to the overall improvement, with the second step being more pronounced, especially on the FCC dataset. 
This is attributed to the longer documents in FCC, with nearly double the number of sentences in each document compared to the other two datasets. This demonstrates that the information provided in Step 1 establishes a solid foundation but is relatively localized. Step 2, involving global information expansion, plays a crucial role in overall enhancement.

To examine the benefits of decomposed execution, we further integrate the two-step workflow into a single-step one through simple concatenation. Despite demonstrating comparable performance on GVC, the integrated workflow shows a noticeable lag, being 1.2\% and 2\% behind on ECB+ and FCC, respectively, in terms of $B^3$ F1. This indicates that even with straightforward instructions, decomposing the multi-objective task into multiple independent steps is necessary, as evidenced by the recent LLM agent studies~\citep{aksitov2023rest}.

We perform error analysis and compare the lengths of the generated summaries to provide a detailed explanation of the impact of each step in the workflow and its decomposition. Further details can be found in Appendix~\ref{sec:workflow_ana}.

\subsection{Analysis of GPT-4 Performance on CDECR}
\label{sec:anal_gpt4}

Table~\ref{tab:eval_gpt} presents the results of different in-context learning methods. It shows that GPT-4 achieves its optimal performance using few-shot learning with mention-inclusive sentences as context (Few-MIS), yet it only achieves results comparable to the lemma matching-based method. 
Table~\ref{tab:gpt_error_stat} further compares different types of errors. Compared to our baseline and our method, Few-MIS has a slight reduction in FPT errors but a significant increase in FPA and FN errors. This indicates that GPT-4 has limited ability to differentiate between similar but non-coreferential events based on arguments, and struggles to link coreferential events with significant narrative differences based on semantics. The reduction in FPT errors may also be attributed to its limited comprehension ability, thereby avoiding errors caused by excessive interpretation of event types. This aligns with our observation that GPT-4 relies on a simplistic approach of clustering based on the literal meaning of mentions without considering their contexts. Additionally, the role of demonstrations appears limited to expanding the scope of matching for synonymous mentions.

From Table~\ref{tab:eval_gpt}, it is also evident that incorporating the full context, compared to solely utilizing mention-inclusive sentences as context, results in a significant performance decline. This indicates that with more context, GPT-4 not only has limited ability to extract effective cues but also suffers from disrupted comprehension of the local context.
Additionally, compared to few-shot learning, zero-shot learning demonstrates higher recall but significantly lower precision. This is because many completely unrelated mentions are clustered into a single cluster. This highlights the complexity of the CDECR task, indicating that GPT-4 struggles to perform basic clustering when relying solely on the task description.

\begin{table}[t]
\centering
\resizebox{0.48\textwidth}{!}{%
\begin{tabular}{llccc}
\hline
\multicolumn{2}{l}{Method}                                         & R    & P    & F1   \\ \hline
\multicolumn{2}{l} {\textsc{Cluster+Lemma}~\cite{barhom-etal-2019-revisiting}} & 71.7 & 85.0 & 77.8 \\
\multicolumn{2}{l} {Our baseline} & 87.1 & 86.0 & 86.5  \\
\multicolumn{2}{l} {Our method} & 89.1 & 86.5 & 87.8 \\ \hline
\multirow{2}{*}{Few-shot}                 & Mention-inclusive sentences               & 76.3 & 78.1 & 77.2 \\
& Full context               & 65.6 & 77.2 & 70.9 \\ \hline
\multirow{2}{*}{Zero-shot}                & Mention-inclusive sentences              & 78.6 & 60.4 & 68.3 \\
& Full context               & 75.6 & 56.4 & 64.6 \\ \hline
\end{tabular}%
}
\caption{Results on ECB+, based on the $B^3$ metric.}
\label{tab:eval_gpt}
\end{table}

\begin{table}[t]
\centering
\resizebox{0.3\textwidth}{!}{%
\begin{tabular}{llccc}
\hline
Methods & FPA & FPT & FN \\ \hline
Our baseline & 1775 & 302 & 1262 \\
Our method & 1227 & 152 & 1087  \\
Few-MIS   & 2272 & 116 & 3435  \\ \hline
\end{tabular}%
}
\caption{Statistics of errors by different types. Few-MIS corresponds to the best-performing in-context learning method from Table~\ref{tab:eval_gpt}, which is few-shot learning with mention-inclusive sentences as context.}
\label{tab:gpt_error_stat}
\end{table}

We further investigate the impact of the number of in-context demonstrations on GPT-4's performance. Details can be found in Appendix ~\ref{sec:demo_num_impact}.

\section{Conclusion}
We design generic tasks to leverage the potential of LLMs for CDECR, effectively bridging the gap between the general capabilities of LLMs and the complex annotation guidelines of specific IE tasks. Results show that by harnessing the inherent knowledge and comprehension abilities of LLMs to gain a deeper understanding of events, our collaborative approach can alleviate the challenge of SLMs for complex contextual understanding, ultimately enhancing performance.

\section*{Limitations}
The LLM we use for our collaborative approach is \texttt{GPT-4-0613}. Moving forward, we plan to assess the performance of additional LLMs, such as LLaMa~\citep{touvron2023llama}. 

For CDECR, where internal information within the given document might be insufficient, there arises a need for external information retrieval. We are considering further leveraging the capabilities of LLMs to explore how to retrieve supplementary information from external corpora such as news articles. Our aim is to combine this additional information with the given documents to enhance performance.

\section*{Ethics Statement}
We adhere to the ACL Code of Ethics.

\section*{Acknowledgement}
We acknowledge the assistance of Pai Liu from University of Rochester in conducting experiments related to GPT-4.
This work was supported by the STI 2030—Major Projects (Grant No. 2022ZD0208800) and the Alibaba Innovative Research (Grant No. 10313H022101).

\bibliography{anthology,custom}

\newpage
\appendix

\section{Implementation details}
\label{sec:appendix}

\subsection{Dataset Statistics}
\label{sec:dataset}
As shown in Table~\ref{tab:dataset_stat}.

\begin{table}[t]
\centering
\resizebox{0.4\textwidth}{!}{%
\begin{tabular}{lccc}
\hline
                  & ECB+  & GVC   & FCC    \\ \hline
Documents         & 982   & 510   & 451    \\
Sentences         & 16314 & 9782  & 14940  \\
Event mentions    & 6833  & 7298  & 3563   \\
Event clusters    & 2741  & 1411  & 469    \\
Event coref links & 26712 & 29398 & 145272 \\ \hline
\end{tabular}%
}
\caption{Statistics of each dataset.}
\label{tab:dataset_stat}
\end{table}

\subsection{SLM Fine-tuning Hyper-parameters}
\label{sec:hyper_param}

As shown in Table~\ref{tab:fine_tune_param}.

\begin{table}[t]
\centering
\resizebox{0.43\textwidth}{!}{%
\begin{tabular}{lcc}
\hline
 & \begin{tabular}[c]{@{}c@{}}Candidate\\ retrieval\end{tabular} & \begin{tabular}[c]{@{}c@{}}Pairwise\\ classification\end{tabular} \\ \hline
Learning rate       & 1e-5 & 6e-6 \\
Batch size          & 16   & 16   \\
Epochs              & 50   & 20   \\
Early stop patience & 10   & 5    \\
Train neighbor size & -    & 20   \\
Eval neighbor size  & 10   & 10   \\ \hline
\end{tabular}%
}
\caption{Hyper-parameters for fine-tuning the SLM-based modules.}
\label{tab:fine_tune_param}
\end{table}

\subsection{ Prompt Design and Model Details of GPT-4 Evaluation}
\label{sec:gpt_prompt}

The prompt is shown in Table~\ref{tab:eval_temp}. For ECB+, we introduce only one randomly selected topic from the training data as the demonstration, which comprises 39 documents, accounting for 6.6\% of the entire training set.
For GVC and FCC, we use the same number of randomly selected documents as in ECB+ for demonstration. 
It is important to note that we have conducted multiple rounds of prompt optimization to ensure GPT-4's performance, including:
\begin{itemize}
\item Designing a reasonable format to tag each mention in the document with a unique \textit{mention\_id} to avoid literal confusion.
\item Designing the output format as \textit{mention\_id: cluster\_index} instead of \textit{cluster\_index: [mention\_id1, ..., mention\_idn]} to ensure that no mention is omitted.
\item Avoiding declaring specific conditions for event coreference in the task description, including coreferential participants, locations, and times. It is demonstrated that these conditions do not improve performance; instead, they lead GPT-4 to make coreference judgments based solely on individual conditions.
\end{itemize}

We set the model parameters, including seed and temperature, to 0 to minimize randomness. Additionally, we specify the output format to be in JSON for better post-processing.

During our experimentation, there were changes in the GPT-4 model provided by OpenAI. The introduction of ``GPT-4-turbo-preview'', which can handle longer texts compared to ``GPT-4-0613'', offers conditions for lenghy context composed by multiple documents (although it still faces length limitations in our actual testing). Consequently, our direct evaluation of the GPT-4 model was moved to ``GPT-4-turbo-preview''.

Since most of our summary-based experiments were completed on ``GPT-4-0613'', we did not migrate our experiments to ``GPT-4-turbo-preview'' due to cost considerations. Additionally, based on our observation with minimal use cases and external leaderboard\footnote{\href{https://huggingface.co/spaces/lmsys/chatbot-arena-leaderboard}{https://huggingface.co/spaces/lmsys/chatbot-arena-leaderboard}}, ``GPT-4-turbo-preview'' (currently pointing to ``GPT-4-0125-preview'') exhibits performance that is not inferior to ``GPT-4-0613''.

\begin{table*}[t]
\centering
\resizebox{\textwidth}{!}{%
\begin{normalsize}
\begin{tabular}{ll}
\hline
System role &
  \begin{tabular}[c]{@{}l@{}}You are a helpful assistant tasked with clustering coreferential event mentions in the provided documents.\\ The event mentions in the documents are marked as follows: [mention string](mention id). Please output\\ the result in JSON format without whitespace. In the JSON structure, each `mention id' is assigned\\ a `cluster id'. \end{tabular}\\ \hline
Prompt &
  \begin{tabular}[c]{@{}l@{}}You can learn from the following example:\\ Input: \\ Document: [\texttt{... [mention](mention\_id) ...}] \\ ... \\ Output: [\texttt{mention\_id: cluster\_id, ...}]\\ Now the following is your task:\\ Document: [\texttt{... [mention](mention\_id) ...}]\\... \end{tabular} \\ \hline
\end{tabular}%
\end{normalsize}
}
\caption{The few-shot prompt for GPT-4 evaluation. The system role is used to declare task requirements and output specifications. The prompt is divided into two sections: initially, a demonstration, followed by data to be processed. For zero-shot, it suffices to remove the demonstration part.}
\label{tab:eval_temp}
\end{table*}

\subsection{Event Type Categorization}
\label{sec:type_cate}

\begin{table}[t]
    \centering
    \resizebox{0.35\textwidth}{!}{%
    \begin{tabular}{cccc}
    \hline
         & Mentions & Clusters & Types\\ \hline
        ECB+ & 1780 & 805 & 405\\ \hline
        GVC & 1008 & 194 & 4\\ \hline
        FCC & 1074 & 167 & 19 \\ \hline
    \end{tabular}
    }
    \caption{Statistics for mention, cluster, and event type in the test set.}
    \label{tab:type_stat}
\end{table}

To categorize event types, we establish a three-layer hierarchical structure of (mention->cluster->type), linking types between mentions. Specifically, if there are synonymous mentions between any two clusters, they belong to the same event type, and all mentions within the clusters belong to a synonymous event type. Drawing inspiration from \cite{ahmed-etal-2023-2}, we determine mention synonymity by matching their span words. Table~\ref{tab:type_stat} illustrates that the contents of FCC and GVC belong to the same topic, resulting in a concentrated set of event types. Conversely, ECB+ involves various topics such as quake, murder, acquisition, etc., leading to a diverse set of event types.

\subsection{LLM Paraphrase Prompt}
\label{sec:para_prompt}
As shown in Table~\ref{tab:para_prompt}.

\begin{table*}[t]
\centering
\resizebox{\textwidth}{!}{%
\fontsize{5}{5.7}\selectfont
\begin{tabular}{l}
\hline
  \begin{tabular}[c]{@{}l@{}}\textbf{News}: \texttt{[input document]}\\ \textbf{Question 1:} In this news, given ``\texttt{[mention 1]}'' mentioned in the sentence ``\texttt{[the sentence]}''.\\ Concatenate the preceding five sentences of the current sentence (ignore if not available), the current sentence,\\ and the subsequent five sentences of the current sentence (ignore if not available) into a single paragraph. Then,\\ paraphrase the concatenated paragraph while preserving the mention \texttt{[mention 1]}. Attempt to express the\\ information differently while maintaining the meaning and key information. Ensure that the mention \texttt{[mention 1]}\\ is preserved and marked as \#\texttt{[mention 1]}\# in the paraphrased result. Limit the paraphrased result to three sentences.\\ Present the information in the following format: `Paraphrase: <placeholder>'. \\ \textbf{Question 2:} ...
  \end{tabular}\\ \hline
\end{tabular}%
}
\caption{The prompt for LLM paraphrase. Each prompt includes a document along with multiple event mentions. 
The content to be filled is represented as [\texttt{content}].} 
\label{tab:para_prompt}
\end{table*}

\section{Experimental Results and Analysis}

\subsection{API Efficiency and Truncation Issues}
\label{sec:effi_trunc}

Compared to directly prompting GPT-4 for structured predictions of event coreference, our two-step prompting for summarizing each document's mentions does incur more API calls and token consumption, as shown in Table~\ref{tab:consum_stat}. The primary additional overhead comes from generating more natural summaries for each mention rather than a final cluster label, which is the core of our collaborative approach.

\begin{table}[t]
\centering
\resizebox{\columnwidth}{!}{%
\begin{tabular}{lccc}
\hline
\multicolumn{1}{l}{} & \multicolumn{1}{l}{Input tokens} & \multicolumn{1}{l}{Output tokens} & \multicolumn{1}{l}{API calls} \\ \hline
Directly prompting & 166k+ & 25k+ & 10 \\ \hline
Our two-step prompting & 658k+ & 213k+ & 400+ \\ \hline
\end{tabular}%
}
\caption{Statistics of token consumption and API calls on ECB+ test set.}
\label{tab:consum_stat}
\end{table}

Based on our approach, we also strive to enhance the efficiency of GPT-4 utilization, including:

\begin{itemize}
    \item Processing all mentions within the same document simultaneously: this avoids assigning a separate document input for each mention, thereby reducing the number of API calls and token consumption, thus improving efficiency. To ensure the accuracy of parallel processing, we employ a concise pre-step (e.g., dependency parsing) integrated into step 1, as described in Section~\ref{sec:setting}.
    \item We strive to summarize event mentions through designing concise prompts, thereby avoiding the additional comsumption of complex inference chains and in-context learning methods.
    \item Some recent work aims to improve the performance of LLMs by having them generate complex reasoning logic, as mentioned in the part of \textbf{Integration of LLM and SLM} in related work. This approach typically involves dealing with a large number of combinations of mention pairs . In comparison, our collaborative approach only requires processing each document's mentions once, thus offering a relative efficiency advantage while enhancing performance.
\end{itemize}

We can also illustrate the truncation issue through the statistics in Table~\ref{tab:consum_stat}. It shows that processing all test data from ECB+ consumed over 166k tokens for input and 25k tokens for output. With GPT-4's output limited to 4096 tokens per instance, processing all test data in one go would allow us to get results for only 15\% of the total mentions. The similar issue primarily results in poor performance on the GVC and FCC datasets. In the future, we will explore ways to address the length issues caused by multi-document scenarios, possibly through multiple processing iterations.

\subsection{Evaluation Under the Conditions of Without Singletons and at the Topic Level}
\label{sec:eval_singleton}

\begin{table*}[ht]
\centering
\resizebox{\textwidth}{!}{%
\begin{tabular}{llccccccccccccc}
\hline
\multicolumn{2}{l}{\multirow{2}{*}{Methods}} & \multicolumn{3}{c}{MUC} & \multicolumn{3}{c}{$B^3$} & \multicolumn{3}{c}{CEAF$_e$} & \multicolumn{1}{c}{CoNLL} & \multicolumn{3}{c}{LEA} \\ \cline{3-15} 
\multicolumn{2}{l}{} & R & P & F1 & R & P & F1 & R & P & F1 & F1 & R & P & F1 \\ \hline
\multirow{2}{*}{\citet{cattan2021realistic}} & singleton+ & 85.1 & 81.9 & 83.5 & 82.1 & 82.7 & 82.4 & 75.2 & 78.9 & 77.0 & 81.0 & - & - & - \\
 & singleton- & 85.1 & 81.9 & 83.5 & 70.8 & 70.2 & 70.5 & 68.2 & 52.3 & 59.2 & 71.1 & - & - & - \\ \hline
\multirow{2}{*}{\citet{chen-etal-2023-cross}} & singleton+ & 88.6 & 85.9 & 87.2 & 87.8 & 85.4 & 86.6 & 82.8 & 83.7 & 83.2 & 85.7 & - & - & - \\
 & singleton- & 88.6 & 85.9 & 87.2 & 76.1 & 74.5 & 75.3 & 76.9 & 57.4 & 65.7 & 76.4 & - & - & - \\ \hline
\multirow{2}{*}{Our baseline} & singleton+ & 86.6 & 86.8 & 86.7 & 87.1 & 86.0 & 86.5 & 82.6 & 82.5 & 82.5 & 85.2 & 77.8 & 76.6 & 77.2 \\
 & singleton- & 86.6 & 86.8 & 86.7 & 80.9 & 77.0 & 78.9 & 69.5 & 62.9 & 66.0 & 77.2 & 77.1 & 71.2 & 74.0 \\ \hline
\multirow{2}{*}{Our method} & singleton+ & 89.4 & 87.1 & 88.2 & 89.1 & 86.5 & 87.8 & 82.7 & 85.5 & 84.1 & 86.7 & 79.7 & 78.5 & 79.3 \\
 & singleton- & 89.4 & 87.1 & 88.2 & 84.0 & 79.9 & 81.9 & 75.3 & 64.9 & 69.7 & 79.9 & 80.9 & 73.9 & 77.2 \\ \hline
\end{tabular}%
}
\caption{Performance comparison on the ECB+ dataset with(singletons+)/without(singletons-) singletons. We are the first to present results under the LEA metric.}
\label{tab:rm_sing_result}
\end{table*}

The experimental results under the condition of with/without singletons are presented in Table~\ref{tab:rm_sing_result}. 
The results demonstrate that our method achieves state-of-the-art performance, surpassing \citet{chen-etal-2023-cross} by 3.5\% and our baseline by 2.7\% in CoNLL F1 under the without singletons condition. Additionally, our method demonstrates a relatively smaller performance gap between with and without singletons compared to \citet{chen-etal-2023-cross} (6.8\% vs 9.3\%) and our baseline (6.8\% vs 8.0\%). This further emphasizes the effectiveness of our method.

For topic level evaluation, it advocates not using subtopic-level document clustering, forcing models to confront the lexical ambiguity challenge. Our method, based on a baseline that performs candidate coreferential mention retrieval at a global level, avoids leveraging topic structure information and achieves better results than subtopic clustering methods. Therefore, we do not perform additional comparisons at the topic level.

\subsection{False Negative Cases}
\label{sec:fn_instances}

Given the context where mentions of the same event can vary greatly in expression styles, we provide an illustrative example in Table~\ref{tab:varied_expression}.

\begin{table}[ht]
\centering
\resizebox{\columnwidth}{!}{%
\begin{tabular}{lc}
\hline
\textbf{Event} & Smith case as the incarnation of the Doctor \\ \hline
\multirow{3}{*}{\begin{tabular}[c]{@{}l@{}}\textbf{Mention} \\ \textbf{expressions}\end{tabular}} & was handed the keys to the Tardis \\
      & winning the role of the 11th Doctor         \\
      & stepping into Doctor Who's title role       \\ \hline
\end{tabular}%
}
\caption{Variations in mention expressions for identical event.}
\label{tab:varied_expression}
\end{table}

In cases where event mentions naturally lack sufficient details, we illustrate this phenomenon through Table~\ref{tab:coref_examples}, which presents two mentions of the same earthquake. The context for the first mention contains essential information such as time, location, magnitude, casualties, etc. In contrast, the context for the second mention primarily describes the subjective experiences of the individuals involved, lacking details related to the event itself. Despite our summarization extracting key information from the original context, it encounters difficulties in supporting coreference judgments.

\begin{table*}[ht]
\centering
\resizebox{\textwidth}{!}{
\fontsize{10}{12}\selectfont
\begin{tabular}{p{0.56\textwidth}p{0.44\textwidth}}
\hline
\multicolumn{1}{l}{Context} &
\multicolumn{1}{l}{Summarization} \\ \hline
[\textit{4.6 earthquake}] shakes Northern California March 14, 2013 | 8:16 am A magnitude 4.6 earthquake struck Northern California on Thursday morning, The temblor struck about 26 miles north of Santa Rosa in the Geysers area. There were no reports of damage or injuries, but there were some apparent aftershocks. It was felt over a wide area of the region, including in Santa Rosa, Marin County and Vallejo, according to the U.S. Geological Survey. The "Do You Feel It?" survey by the USGS showed people as far south a San Francisco felt shaking.  &
[\textit{4.6 earthquake}] refers to the seismic event that occurred in \textbf{Northern California} on \textbf{March 14, 2013}, with a magnitude of 4.6 on the Richter scale. The earthquake was felt over a wide area of the region, including in \textbf{Santa Rosa, Marin County, and Vallejo}, and caused some apparent aftershocks. However, there were no reports of damage or injuries. \\ \hline
Good sized quake at the California Geysers I felt this one pretty good. I'm located 25 miles south of the quake. Slow easy rocking for about 6 seconds. This one is pretty rare for the Geysers. The Geysers is one of the most active areas of Northern California with small earthquakes hitting the area nearly every week. A 4.4 [\textit{quake}] is on the large size for The Geysers. The UGSG Website says the largest quake ever to be recorded there was 4.5. &
[\textit{quake}] refers to a seismic event that occurred at the \textbf{California Geysers}, which was felt \textbf{25 miles south of the epicenter}. The quake was described as a slow, easy rocking that lasted for about 6 seconds. The \textbf{exact date of the quake is not provided} in the news, but it can be inferred to have occurred recently based on the publication date of the news. \\ \hline
\end{tabular}
}
\caption{Two coreferential mentions referring to the same earthquake, where the second provides minimal coreference evidence. Key information in our summarization is highlighted in bold. Mention spans are represented as [\textit{mention span}]. }
\label{tab:coref_examples}
\end{table*}

\subsection{Two-step Workflow Analysis}
\label{sec:workflow_ana}

\paragraph{Error Analysis} We conduct error analysis for the workflow with only Step 1, the complete two-step workflow (Step 2), and the integrated single-step workflow. 

\begin{figure*}[ht]
    \centering
    \includegraphics[width=1\linewidth]{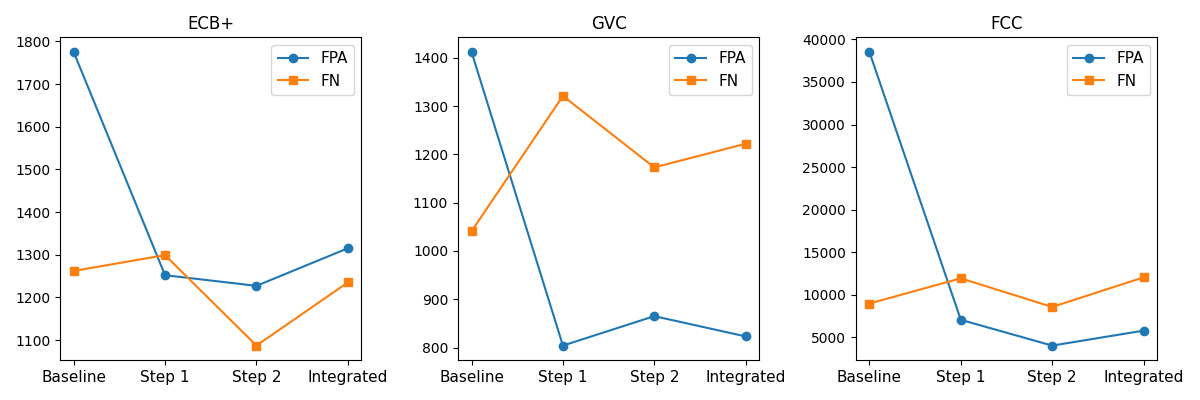}
    \caption{FPA and FN error comparison. Due to the rarity of FPT-type errors, we have omitted them in the figures for better clarity in presentation. Step 2 is built upon Step 1, and the integrated involves merging the two steps together.}
    \label{fig:workflow_error}
\end{figure*}

As shown in Figure~\ref{fig:workflow_error}, Step 1 exhibits a significant reduction in FPA errors across all three datasets, indicating its effectiveness in extracting tailored information. However, an increase in FN errors is observed across all three datasets, suggesting that while Step 1 provides sufficiently distinctive information, it lacks the details needed to link mentions of the same event. This issue was notably addressed by the introduction of Step 2, resulting in a substantial decrease in FN errors across all datasets. FPA errors are also largely maintained at the level achieved in Step 1, leading to a significant improvement in coreference results. This emphasizes the indispensable roles of both Step 1 and Step 2 in the final outcomes. In Table~\ref{tab:workflow_examples}, we provide examples to compare summaries generated by Step 1 and Step 2. 

Compared to the two-step workflow, the integrated single-step workflow shows differing degrees of increase in both FPA and FN errors, further underscoring the necessity of decomposed execution.

\begin{table*}[t]
\centering
\resizebox{\textwidth}{!}{
\fontsize{10}{12}\selectfont
\begin{tabular}{p{0.3\textwidth}p{0.7\textwidth}}
\hline
\multicolumn{1}{l}{Step 1} &
\multicolumn{1}{l}{Step 2} \\ \hline
[\textit{6.1-magnitude earthquake}] refers to the seismic event that occurred in Aceh, Indonesia, with a magnitude of 6.1 on the Richter scale. &
[\textit{6.1-magnitude earthquake}] refers to the seismic event that occurred in the \textbf{Bener Meriah district in the heart of Aceh}, Indonesia, on \textbf{July 2, 2013}. The earthquake struck inland at 0737 GMT at a depth of just 10 kilometres (6.2 miles) and was felt strongly for around 15 seconds, \textbf{from Bener Meriah to Banda Aceh}. \\ \hline
[\textit{earthquake}] refers to the magnitude 6.1 earthquake that hit Indonesia's West Papua province.  &
[\textit{earthquake}] refers to the magnitude 6.1 earthquake that hit Indonesia's \textbf{West Papua} province on an \textbf{unspecified date}. The earthquake struck off the coast at 7:48 a.m. local time, 75 kilometers (50 miles) west of the region's \textbf{main city of Manokwari}, according to the U.S. Geological Survey. \\ \hline
\end{tabular}
}
\caption{Comparison of summaries generated by Step 1 and Step 2. Step 2 is built upon Step 1. Key information for distinguishing in Step 2 is highlighted in bold. Mention spans are represented as [\textit{mention span}].}
\label{tab:workflow_examples}
\end{table*}

\begin{figure*}[ht]
    \centering
    \includegraphics[width=1\linewidth]{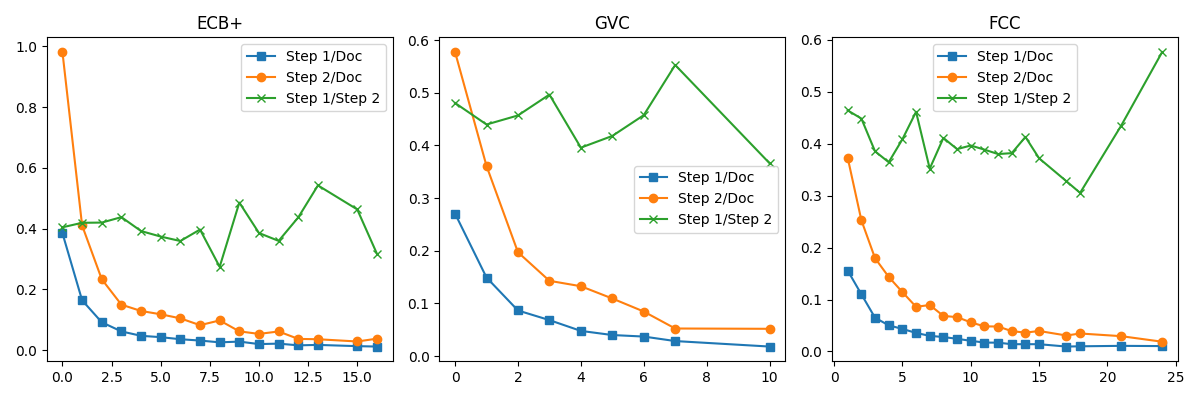}
    \caption{Summarization length comparison. Step 2 is built upon Step 1. The vertical axis represents the ratio of content word count. The horizontal axis represents the number of words in the content, scaled by a factor of 100.}
    \label{fig:len_ratio}
\end{figure*}

\paragraph{Summarization length comparison}
We further compare the lengths of summaries generated in Step 1 and Step 2. As illustrated by the green line in Figure~\ref{fig:len_ratio}, it is evident that Step 2, building upon Step 1, results in approximately double the length. The additional detailed content contributes to the reduction of FN errors, effectively linking mentions of the same event. Furthermore, as indicated by the red and blue lines,  our generated summaries remain within approximately 20\% of the original document starting from a document length of 200 words. Moreover, with the increase in document length, this proportion further diminishes. This reflects the conciseness our summarization.

\subsection{The Impact of the Number of In-Context Demonstrations on GPT-4 Performance on CDECR}
\label{sec:demo_num_impact}
We test the peak performance by increasing the number of documents for demonstration. The results are shown in Figure~\ref{fig:demo_num_impact}, and it can be observed that:
\begin{itemize}

    \item Under the condition of utilizing only mention-inclusive sentences as context, with the introduction of more documents (even exceeding the quantity in the test set), there is still no significant improvement in the performance of GPT-4. And there remains a considerable gap compared to the F1 score of our method (77.2\% vs 87.8\%).
    \item Under the condition of utilizing full context, an increase in the number of documents can even degrade performance. Since the complete context is crucial for event coreference resolution, it indicates that understanding and utilizing more context is a significant bottleneck limiting the performance of GPT-4.

\begin{figure}[t]
    \centering
    \includegraphics[width=1\linewidth]{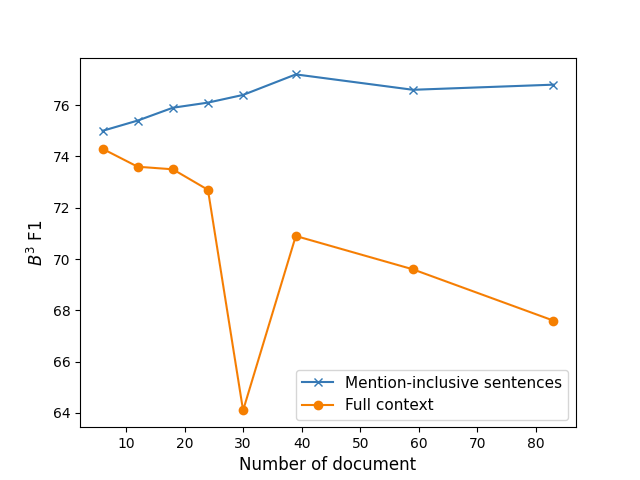}
    \caption{The impact of number of demonstrations on GPT-4 performance, measured by controlling the number of documents used. In our main experiments evaluating GPT-4, we utilize one instance of demonstration comprising 39 documents.}
    \label{fig:demo_num_impact}
\end{figure}

\end{itemize}

\end{document}